\documentclass[journal,10pt]{IEEEtran}
\ifCLASSINFOpdf
\else
\fi
\usepackage{microtype}
\usepackage{graphicx}
\usepackage{subfigure}
\usepackage{booktabs} 
\usepackage{amssymb,amsmath}

\usepackage{color}
\usepackage[english]{babel}
\usepackage{array}
\usepackage{bm}
\usepackage{multirow}
\usepackage{enumitem}
\usepackage{pifont}
\usepackage{tikz}
\usepackage{amssymb}
\usepackage{enumitem}
\usepackage{cite}
\usepackage[english]{babel}
\usepackage{array}
\usepackage{bm}
\usepackage{multirow}
\usepackage{enumitem}
\usepackage{pifont}
\usepackage{flushend}

\newcommand{\revise}[1]{\textcolor[rgb]{0.00,0.00,0.00}{#1}}
\newcommand{\rtypos}[1]{\textcolor[rgb]{0.00,0.00,0.00}{#1}}

\newtheorem{theorem}{Theorem}

\newtheorem{property}[theorem]{Property}

\def\placeholder{HotNAS}

\begin{document}
\title{Standing on the Shoulders of Giants: Hardware and Neural Architecture Co-Search with Hot Start}

\author{Weiwen Jiang,
        Lei Yang,
        Sakyasingha Dasgupta,
        Jingtong Hu,
        and~Yiyu Shi
\thanks{
  W. Jiang and Y. Shi are with the Department of Computer Science and Engineering, University of Notre Dame, Notre Dame, IN 46556.
}
\thanks{L. Yang is with the Department of Electrical and Computer Engineering, University of New Mexico, Albuquerque, NM 87131.
}
\thanks{S. Dasgupta is with Edgecortix Inc., Tokyo, Japan, 1410031.
}
\thanks{J. Hu is with the Department of Electrical and Computer Engineering, University of Pittsburgh, Pittsburgh, PA 15261.
}
\thanks{W. Jiang is the corresponding author (e-mail: wjiang2@nd.edu).
}
\thanks{This work is partially supported by the National Science Foundation under Grants CCF-1919167, CCF-1820537 and CNS-1822099. 
}
\thanks{This article was presented in the International Conference on Hardware/Software Codesign and System Synthesis 2020 and appears as part of the ESWEEK-TCAD special issue.
}
}

\markboth{IEEE Transactions on Computer-Aided Design of Integrated Circuits and Systems}
{Jiang \MakeLowercase{\textit{et al.}}: Standing on the Shoulders of Giants: Hardware and Neural Architecture Co-Search with Hot Start}

\maketitle

\begin{abstract}
Hardware and neural architecture co-search that automatically generates Artificial Intelligence (AI) solutions from a given dataset is promising to promote AI democratization; however, \rtypos{the amount of time that is required by current co-search frameworks is in the order of hundreds of GPU hours for one target hardware. This inhibits the use of such frameworks on commodity hardware.} 
The root cause of the low efficiency \rtypos{in existing} co-search frameworks is \rtypos{the fact that they start from a ``cold'' state} (i.e., search from scratch).
In this paper, we propose a novel framework, namely \placeholder, that starts from \rtypos{a ``hot'' state based on} a set of existing pre-trained models (\rtypos{a.k.a.} model zoo) to avoid lengthy training \rtypos{time}.
As such, the search time can be reduced from 200 GPU hours to less than 3 GPU hours.
In \placeholder, in addition to hardware design space and neural architecture search space, we further \rtypos{integrate} a compression space to conduct model compressing during the co-search, which \rtypos{creates new} opportunities to reduce latency, but also brings challenges.
One of the key challenges is that all \rtypos{of} the above search spaces are coupled with each other, e.g., compression may not work without hardware design support.
\rtypos{To tackle this issue}, \placeholder~builds a chain of tools to design hardware to support compression, based on which a global optimizer is developed to automatically co-search all the involved search spaces. 
Experiments on \rtypos{ImageNet dataset} and Xilinx FPGA show that, within the timing constraint of 5ms, neural architectures generated by \placeholder~can achieve up to 5.79\% Top-1 and 3.97\% Top-5 accuracy gain, compared with the existing ones.


\end{abstract}

\setlength{\textfloatsep}{3pt}
\setlength{\floatsep}{3pt}
\setlength{\dbltextfloatsep}{3pt}

%
\IEEEpeerreviewmaketitle

\section{Introduction}



{The success of Deep Neural Networks (DNN), has propelled Artificial Intelligence (AI) in entering every aspect of our lives and is being widely employed for diverse applications on different types of hardware.}
Neural Architecture Search (NAS), a successful product of Automatic Machine Learning (AutoML), has paved the way from a given dataset to a neural architecture with state-of-the-art accuracy.
\revise{Moving forward, to be able to use AI for enabling and accelerating different applications, we need to be able to design the neural network in a way that the design specifications are met on our target hardware}; for instance, real-time constraints for edge devices, low power budgets for IoT devices, etc.

Recently, neural architecture and hardware design \rtypos{(abbr. architecture-hardware)} co-search frameworks \cite{wu2019fbnet,cai2018proxylessnas,tan2019mnasnet,hao2019fpga,jiang2019accuracy,jiang2020hardware,jiang2020device,yang2020co,yang2020coasp,zhang2019bi,li2020edd,bian2020nass,jiang2016optimal} \rtypos{have been} proposed \rtypos{to bridge the gap between neural architecture and hardware design.}
These frameworks have demonstrated promising results in generating high-accuracy and low-cost systems. \rtypos{However,} their search efficiency is low:  existing co-search frameworks commonly take hundreds of GPU hours \rtypos{per target hardware.} This may become the bottleneck in many emerging applications where fast turn-around or short time-to-market is desired.
On the other hand, it has already been shown that the carbon footprint (\rtypos{pounds} of $CO_2$) of NAS for one model is nearly equivalent to five times the lifetime emissions of a car \cite{strubell2019energy}.
\rtypos{In this work}, we are revisiting \rtypos{the} default setting used by existing co-search frameworks, \rtypos{where}: the exploration always starts from scratch (i.e., cold start), which \rtypos{results in large search time and low efficiency}. \rtypos{However,} is \rtypos{a cold start} really necessary?    

\begin{figure}[t]
  \centering
  \includegraphics[width=0.9\linewidth]{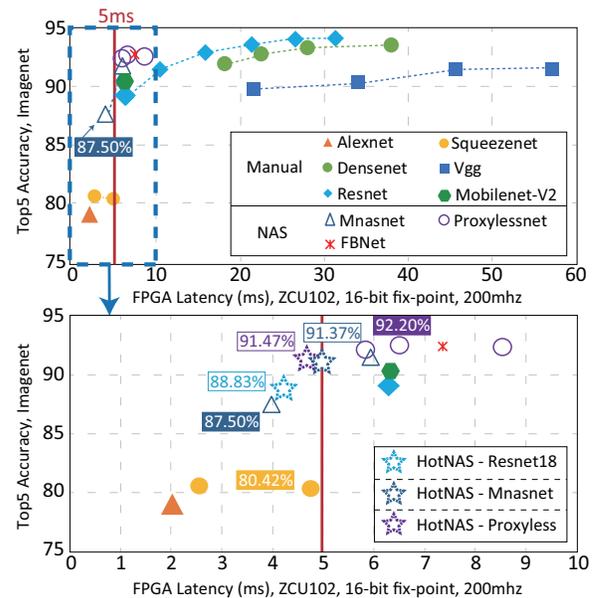}
  \caption{Architecture-hardware co-search by \placeholder: 
  (top) \rtypos{latency and accuracy of models in the} model zoo; \rtypos{(bottom) architectures in model zoo and that identified by \placeholder with 5ms timing constraint}~(Best viewed in color)}
  \label{fig:existingnets}
\end{figure} 


We claim that the \rtypos{architecture-hardware co-search could} stand on the shoulders of giants \rtypos{and start the search} from a hot \rtypos{state}, i.e., \rtypos{using} an existing pre-trained model in a model zoo.
\revise{The model zoo can be efficiently created, consisting of the existing neural architectures manually designed by domain experts, identified by NAS, or transferred from models from different datasets.} 
To make full use of the candidates in the model zoo, in this paper, we propose a novel co-search framework, namely ``\placeholder'', to \rtypos{start searching} from a hot \rtypos{state}. 
\rtypos{In this way, compared with the cold-start co-search, \placeholder~can \rtypos{reduce} the search time from hundreds of GPU hours to \textbf{less than 3 GPU hours for ImageNet and 20 GPU minutes for CIFAR-10 without proxy}; \rtypos{while achieving accuracy} comparable with the state-of-the-art models.}

Figure \ref{fig:existingnets} shows the results of co-search using a model zoo with 24 models on ImageNet dataset, targeting a system with 5ms on Xilinx ZCU 102 FPGA.
From the top figure, there are only 4 models that can satisfy the timing constraint and the highest accuracy is 87.50\%; however, within the range from 5ms to 10ms, there are a lot of good candidates with accuracy higher than 90\%.
The existing co-search frameworks ignore these candidates and search from scratch, leading to hundreds of GPU hours.
Viewing from the opposite angle, \placeholder~takes full \rtypos{use} of these pre-trained models and customize the models that violate time constraints but have high accuracy to the target hardware.
As such, \placeholder~can avoid lengthy training procedure to generate the solution in a couple of hours, which is guaranteed to meet timing constraints while greatly improving accuracy to 91.47\%, as shown in the bottom figure.



%
\revise{Seemingly straightforward, the architecture-hardware co-search from a hot start is not a simple matter:
a fundamental challenge is the discovery of the best search space.
Some of the prior co-search works \cite{jiang2019accuracy,jiang2020hardware,hao2019fpga,lu2019neural} consider hardware design space of loop tiling and loop order, and neural architecture search space with flexibility across the number of channels, filter size, and model quantization. However, we observe that one of the most efficient techniques, model pruning \cite{han2015deep,mao2017exploring,wen2016learning,ma2019pconv,yawen2020intermittent}, has hitherto not been combined in the co-search.
}
\revise{Integrating model pruning faces a lot of challenges:}
First, without the full consideration of hardware design, the model pruning can easily become useless since it introduces overheads.
Second, one compression technique does not work for all performance bottlenecks.
\revise{Finally, the model compression techniques are tightly coupled with hardware design and neural architecture search: As such, a difficult challenge is to simultaneously optimize all these spaces.}

In \placeholder, we address the above challenges by \rtypos{collaboration among four sub-components: iSearch, iSpace, iDesign, and iDetect}.
First, iDesign provides hardware design support for different compression techniques.
Second, \rtypos{following} the observation that different pruning techniques work for different types of bottlenecks;
iDetect is developed to identify performance bottleneck for each layer so that we can select the most suitable \rtypos{compression} techniques to alleviate \rtypos{performance} bottlenecks. 
According to the detected bottlenecks, iSpace \rtypos{creates} a dedicated search space for each layer.
Finally, iSearch is devised to jointly search the hardware, neural architecture, and model compression \rtypos{using specification from iSpace}.

The main contributions \rtypos{of} this paper are threefold:
\begin{itemize}[noitemsep,topsep=0pt,parsep=0pt,partopsep=0pt]
  \item We propose a \rtypos{novel} neural architecture search mechanism to search from a hot \rtypos{state} (i.e., \rtypos{a} pre-trained model), \rtypos{which allows us to reduce search time} from 200 GPU Hours to 3 GPU Hours; meanwhile, the solution can \rtypos{improve the Top-1 and Top-5 accuracy on the ImageNet dataset by 5.79\% and 3.97\%, respectively.}
  \item An \rtypos{automated} \placeholder~framework is proposed to link the hardware design, neural architecture search, and model compression to automatically generate the architecture and hardware pair, \rtypos{such that the timing constraint can be met with the maximum} model accuracy.
  \item In \placeholder, dedicated hardware designs to support the existing model compression are proposed, without which the model compression techniques may not achieve \rtypos{any performance} gain at all. 
\end{itemize}


The remainder of the paper is organized as follows. Section 2 presents design challenges and motivation. Section 3 presents the proposed \placeholder.
Experimental results are shown in Section 4.
Finally, concluding remarks are given in Section 5.

\section{Challenges and Motivation}

\begin{figure}[t]
  \centering
  \includegraphics[width=1.02\linewidth]{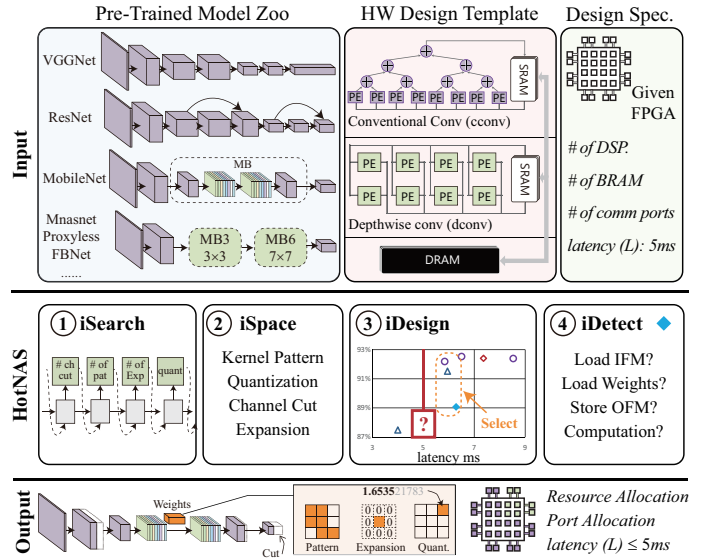}
  \vspace{-10pt}
  \caption{Implementation from the model zoo to hardware: (top) the given pre-trained model and design spec.; (middle) four components in the proposed \placeholder~framework; (bottom) outputting the best neural architecture and hardware design.}
  \label{fig:iNas}
\end{figure}

\rtypos{This section demonstrates the challenges in the architecture-hardware co-search, and gives the motivation of this work.}


Figure \ref{fig:iNas} demonstrates the \rtypos{architecture-hardware} co-search problem, where we have a set of pre-trained models \rtypos{(called model zoo)}, the hardware design templates and the design specifications (e.g., constraints on the resource, area, and latency) as inputs.
\rtypos{The co-search is to optimize neural architectures in the model zoo and hardware design to guarantee all design constraints to be met while maximizing accuracy, as shown in the bottom part in Figure \ref{fig:iNas}} 


\placeholder~framework is proposed in this paper to solve the above problem.
As illustrated in the middle part of Figure \ref{fig:iNas}, it is composed of four sub-components: \textbf{\emph{\raisebox{-1pt}{\large\ding{192}}}} iSearch,  starting the co-search from hot instead of cold;
\textbf{\emph{\raisebox{-1pt}{\large\ding{193}}}} iSpace, building \rtypos{an integrated search space} which is in accordance with the performance bottleneck in the implementation; \textbf{\emph{\raisebox{-1pt}{\large\ding{194}}}} iDesign, \rtypos{providing the design to support compression techniques on FPGAs}; \textbf{\emph{\raisebox{-1pt}{\large\ding{195}}}} iDetect, detecting the performance bottleneck \rtypos{to guide the creation of iSpace}.
In the following text, we will show that there \rtypos{exists} a couple of challenges in \rtypos{architecture-hardware co-search} and all components work collaboratively to address these challenges.


\vspace{5pt}
\noindent\textit{\textbf{Challenge 1:} How to efficiently explore neural architectures.}
\vspace{5pt}

\rtypos{The order of hundreds of GPU hours in architecture-hardware co-search cannot satisfy the short time-to-market requirements in many applications;
as reported in Table \ref{tab:mot_ex2}, the state-of-the-art {hardware agnostic} neural architecture search techniques (DARTS \cite{liu2018darts}) requires 90 GPU Hours, while the NAS for a specific type of hardware (MnasNet \cite{tan2019mnasnet}, FNAS \cite{jiang2019accuracy}, FBNet \cite{wu2019fbnet}, and ProxylessNAS \cite{cai2018proxylessnas}) requires over 200 GPU Hours.}
Considering that the current computing system is composed of a large variety of hardware, the search process is simply unacceptable.
In addition, the long search time \rtypos{leads to} excessive $CO_2$ emission, which has already been known as a serious problem of \rtypos{existing} NAS {techniques} \cite{strubell2019energy}.

We observe that the long search time in NAS framework is caused by the cold start. 
\rtypos{This leads to} more than 40,000 GPU hours for MnasNet and NASNet to train a large number of potential architectures \rtypos{from scratch, and over 200 GPU hours when the hardware is considered.}
\rtypos{However}, \rtypos{there exists} a large set of pre-trained neural networks.
We \rtypos{revisit} the default configuration in the co-search framework: \rtypos{i.e.} whether it is necessary to  \rtypos{start the exploration} from scratch, which results in low efficiency.
In \textbf{\emph{\raisebox{-1pt}{\large\ding{192}}}} iSearch, we propose to \rtypos{make} full use of the existing models and start the exploration from \rtypos{a} hot \rtypos{state} (e.g., pre-trained models).



\begin{table}[t]
  \centering
  \tabcolsep 5pt
  \renewcommand\arraystretch{1.3}
  \scriptsize
  \caption{Search cost (GPU hours) of neural architecture search is too high. Note that the hardware-aware approach needs an entire search for each specific hardware}
    \begin{tabular}{c|cc|cccc}
    \toprule
    NAS   & NASNet & DARTS & MnasNet & FNAS  & FBNet & ProxylessNAS \\
    \midrule
    GPU Hours & 48,000 & 90    & 40,000 & 267   & 216   & 200 \\
    \bottomrule
    \end{tabular}%
  \label{tab:mot_ex2}%
\end{table}%




\vspace{5pt}
\noindent\textit{\textbf{Challenge 2:} Meet real-time constraint on specific hardware.}
\vspace{5pt}

\rtypos{Arbitrarily picking neural networks from the model zoo and plugging into the given hardware will lead to violations of the design specification, e.g., missing deadline in real-time systems.
On the other hand, }due to the large variety of hardware (different types of CPU, GPU, FPGA), it is \rtypos{infeasible} to conduct co-search for all off-the-shelf hardware in advance.


Table \ref{tab:mot_ex1} reports the hardware optimization results for the existing \rtypos{model zoo (Figure \ref{fig:existingnets}(a)) for Xilinx ZCU102 FPGA board with latency requirements of $\le 5ms$.
We can see that }
only four models can satisfy the timing constraints with the highest accuracy of 87.50\%\rtypos{; while} there are a \rtypos{group} of networks whose accuracy is much higher, up to 92.54\%, \rtypos{with the latency slightly exceeding the timing constraint.}

The challenge here is, how can we compress the models to satisfy the timing constraints using its pre-trained weights\rtypos{, while a competitive model accuracy can be achieved.}
\textbf{\emph{\raisebox{-1pt}{\large\ding{193}}}} iSpace is developed to \rtypos{involve model compression in the search space, together with the hardware design space and neural architectures search space.}

\vspace{5pt}
\noindent\textit{\textbf{Challenge 3:} One technique is not for all.}
\vspace{5pt}

\rtypos{One compression technique cannot solve all kinds of performance bottleneck in the accelerator.}
To effectively \rtypos{reduce latency}, we need to specify where the bottleneck is and apply the most suitable compression technique \rtypos{to alleviate} it.


Table \ref{tab:mot_ex1} reports the performance bottleneck analysis.
In the table, column ``Comm. Des.'' indicates the communication subsystem design, where $I_p$, $O_p$, and $W_p$ are communication ports allocated for input feature maps (IFM), output feature maps (OFM) and Weights, respectively.
\rtypos{The total communication bandwidth is limited,}
as a result, the performance of a layer may be bounded by \rtypos{the} specific type of data \rtypos{movement}.
\rtypos{Using this information}, the performance bottleneck can be \rtypos{computed}.
Under the column ``Bottleneck'', \rtypos{the above 4 types of bottlenecks are denoted by I, W, O, C.}
From the table, \rtypos{we can clearly observe} that layers in one network \rtypos{lead} to different types of \rtypos{bottlenecks}.

In \placeholder, \textbf{\emph{\raisebox{-1pt}{\large\ding{194}}}} iDesign will \rtypos{devise hardware} design to support compression \rtypos{techniques}, and \rtypos{provide the performance model; while} \textbf{\emph{\raisebox{-1pt}{\large\ding{195}}}} iDetect will \rtypos{match the compression technique with the corresponding type of performance bottleneck.}


\begin{table}[t]
  \centering
  \tabcolsep 3pt
  \renewcommand\arraystretch{1.3}
  \scriptsize
  \caption{Comparison of the design, bottlenecks, latency, and accuracy of different networks (latency $\le$10ms) on Xilinx ZCU102 board using 16-bit fix-point at 200MHz of frequency. Targeting $5ms$ for ImageNet dataset. (ordered by latency)}
    \begin{tabular}{c|ccc|ccc|cccc|cc}
    \toprule
    \multirow{2}{*}{Network} & \multicolumn{3}{c|}{Comp. Des.} & \multicolumn{3}{c|}{Comm. Des.} & \multicolumn{4}{c|}{Bottleneck} & Lat.  & Top-5 \\
          & Tm    & Tn    & Tm(d) & $I_p$  & $O_p$  & $W_p$  & C & I   & W & O   & (ms)  &  (\%) \\
    \midrule
    alexnet & 70    & 36    & -     & 18    & 6     & 2     & 4     & 0     & 1     & 0     & 2.02  & 79.066 \\
    squeezenet 1.1 & 70    & 36    & -     & 6     & 14    & 10    & 5     & 9     & 4     & 8     & 2.47  & 80.624 \\
    mnasnet 0.5 & 100   & 16    & 832   & 6     & 10    & 14    & 8     & 16    & 16    & 12    & 3.99  & \textbf{87.498} \\
    squeezenet 1.0 & 130   & 19    & -     & 6     & 10    & 14    & 8     & 7     & 4     & 7     & {4.61}  & {80.422} \\
    \midrule
    proxyless mobile & 100   & 16    & 832   & 10    & 10    & 10    & 17    & 15    & 11    & 18    & {5.83}  & {92.202} \\
    mnasnet1.0 & 100   & 18    & 704   & 10    & 10    & 10    & 7     & 21    & 10    & 14    & 5.94  & 91.51 \\
    resnet18 & 70    & 36    & -     & 18    & 6     & 6     & 8     & 2     & 10    & 0     & 6.27  & 89.078 \\
    mobilenet v2 & 160   & 12    & 576   & 10    & 10    & 10    & 2     & 20    & 8     & 22    & 6.29  & 90.286 \\
    proxyless gpu & 130   & 12    & 832   & 10    & 10    & 10    & 8     & 15    & 11    & 9     & 6.49  & \textbf{92.538} \\
    fbnet & 100   & 16    & 832   & 10    & 10    & 10    & 11    & 29    & 10    & 15    & 7.37  & 92.386 \\
    proxyless cpu & 220   & 8     & 704   & 10    & 10    & 10    & 18    & 5     & 10    & 28    & 8.53  & 92.394 \\
    \bottomrule
    \end{tabular}%
  \label{tab:mot_ex1}%
\end{table}%

\section{Proposed Framework: \placeholder}
\label{sec:hotnas}
In response to all the challenges described in the previous section, we propose \placeholder~framework in this section. 
As shown in Figure \ref{fig:iNas}, \placeholder~is composed of four \rtypos{sub-components}, \textbf{\emph{\raisebox{-1pt}{\large\ding{192}}}} iSearch,
\textbf{\emph{\raisebox{-1pt}{\large\ding{193}}}} iSpace, \textbf{\emph{\raisebox{-1pt}{\large\ding{194}}}} iDesign, and \textbf{\emph{\raisebox{-1pt}{\large\ding{195}}}} iDetect.
This section will introduce these \rtypos{sub-components} in detail.

\vspace{5pt}
\noindent{\textbf{\emph{\raisebox{-1pt}{\large\ding{192}}}} \textbf{\textit{iSearch: Search from Hot Start}}}
\vspace{5pt}

\begin{figure}[t]
  \centering
  \includegraphics[width=1\linewidth]{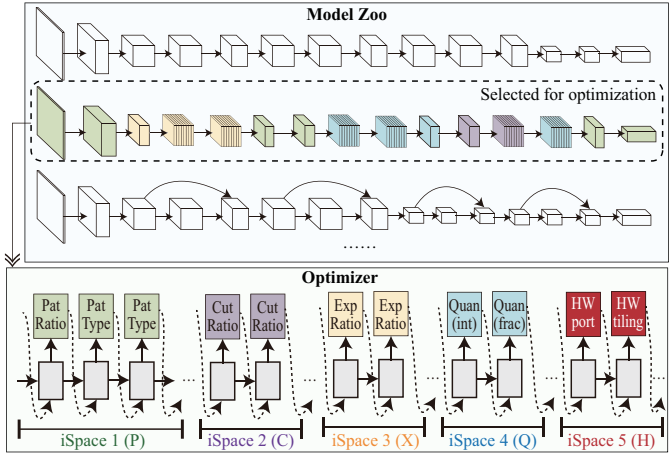}
  \vspace{-10pt}
  \caption{iSearch, iSpace, iDesign, and iDetect: (top) iSearch selects an architecture from \rtypos{the model zoo} for optimization based on the results of \rtypos{iDesign and iDetect}; (bottom) iSpace creates the search space in terms of performance bottlenecks, and \rtypos{iSearch determines hyperparameters for each search space}.}
  \label{fig:iSearch}
\end{figure}

Figure \ref{fig:iSearch} illustrates the overview of iSearch, which conducts the neural architecture search in two steps: (1) \rtypos{(top part of figure)}, it selects backbone architectures to be optimized; then, (2) \rtypos{(bottom part of figure)}, \rtypos{an optimizer tunes} hyperparameters of neural architecture and hardware design simultaneously.
The goal of iSearch is to find the architecture with the highest accuracy while \rtypos{meeting hardware design specifications. In the following texts, we will formally define the problem, and introduce the optimizer at the end of this section.}


\vspace{3pt}
\textit{i) Neural architectures and model zoo:} 
\vspace{3pt}

A neural architecture is defined as $A=\langle V,E,r,c,ch,o,f,para,acc\rangle$, composed of a set of nodes $V$ representing the intermediate data (i.e., input and output feature maps), a set of edges $E\subseteq V\times V$ \rtypos{representing} the dependency between a pair of nodes.
For a node $v_i$ in $V$, it has three hyperparameters $r_i$, $c_i$ and $ch_i$ \rtypos{representing} the number of row, column and channel of $v_i$.
For an edge $e_j\in E$, \rtypos{an operator $o_j$} (e.g., \rtypos{convolution}, depthwise convolution, or pooling, etc.) is associated to it.
$f_j$ represents the filter (i.e., weights) used in operator $o_j$, which is composed of a set of kernels. 
Each filter is associated with two hyperparameters: $s(f_i)$ indicates the size of the filter (e.g., $1\times1$, $3\times3$ etc.), and $p(f_i)$ is \rtypos{a pattern applied to prune $f_i$}.
Both the size and the pattern of the filter are tunable, which will be introduced in \textbf{\emph{\raisebox{-1pt}{\large\ding{193}}}} iSpace.
After all the above hyperparameters are determined \rtypos{and} the neural architecture $A$ is identified, it can be \rtypos{trained/fine-tuned} on the training datasets (e.g., ImageNet) to obtain the parameters/weights $para(A)$, and finally we can obtain its test accuracy $acc(A)$ on the test dataset.   

\rtypos{A pre-trained neural architecture} is also called a model, and a model zoo \rtypos{$M=\{A_0,A_1,\cdots,A_{N-1}\}$} is composed of $N$ models.
\revise{These models can be manually designed by experts, like AlexNet, VGGNet, ResNet, automatically searched via neural architecture search, like MnasNet, ProxylessNas, FBNet, or transferred from models for other datasets, like BiT \cite{Kolesnikov2019BigT}.
In this work, we use the existing model zoo from torchvison, and collect the state-of-the-art pre-trained models from github; hence, the cost of building the model zoo can be neglected.
Kindly note that, how to create the model zoo is out of the scope of this paper; related works can be found in \cite{ying2019bench,Dong2020NAS}.}



\vspace{3pt}
\textit{ii) FPGA and accelerator design:} 
\vspace{3pt}

The hardware efficiency is not only related to the neural architecture but also the resource \rtypos{on} the FPGA and the accelerator design on \rtypos{it}.
An FPGA $fp$ has 3 attributes: $mem_{fp}$, $comp_{fp}$, and $BW_{fp}$, referring to the on-chip memory size, the number of computing resources (e.g., DSPs), and the bandwidth between off-chip and on-chip memories, respectively.

The accelerator design should meet all \rtypos{resource constraints} of a given FPGA.
It is composed of two parts: the design of the computing subsystem and the design of the communication subsystem.
\rtypos{As} the basic operators $o$ in architecture $A$ are conducted in nested loops, the loop optimizations, in particular the loop tiling, are widely studied and used in the design of the computing subsystem in FPGAs \cite{zhang2015optimizing,zhang2016energy}.  
\rtypos{In addition, with the consideration of the large amount of data (i.e., intermediate data and weights), and the limited on-chip buffer in FPGA, it is \rtypos{infeasible} to put all data on FPGA.
Therefore, data are moved between the off-chip and on-chip memories.}
As such, the communication bandwidth for moving \rtypos{each type of data needs to be determined} in the design phase.

As a whole, the accelerator design is defined as $D=\langle T_m,T_n,T_r,T_c, I_b,W_b, O_b\rangle$, containing the loop tiling design $\langle T_m,T_n,T_r,T_c\rangle$ and bandwidth allocation $\langle I_b,O_b,W_b\rangle$.
Specifically, for an operator $o_k$ associated to a pair of nodes $v_i\rightarrow v_j$ in an architecture, \rtypos{$T_m,T_n,T_r,T_c$ are the tiling parameters on output feature maps (OFM) channels $ch_j$, input feature maps (IFM) channels $ch_i$, rows $r_i$, and columns $c_i$; while $\langle I_b,O_b,W_b\rangle$ are the bandwidth allocated for moving IFM (i.e., $v_i$), OFM (i.e., $v_j$), and weights (i.e., $f_k$).}
For a design $D$ and an architecture $A$, the latency of each operator, say $o_k$, can be determined with \textbf{\emph{\raisebox{-1pt}{\large\ding{194}}}} iDesign tool. Then, the summation of all operators will be the latency of $A$, denoted as $lat(A)$.

\vspace{3pt}
\textit{iii) iSearch: two-step exploration} 
\vspace{3pt}

\revise{In iSearch, the first step is to select a set of candidate backbone architectures to be optimized.
Given a neural architecture and an FPGA, it has already been well studied to obtain the best accelerator design $D$, as in \cite{zhang2015optimizing}.
Based on the design, \placeholder~can generate the search space iSpace (Section \ref{sec:hotnas} \textbf{\emph{\raisebox{-1pt}{\large\ding{193}}}}).
iSearch will select models from the model zoo to be the backbone architecture, which will be the starting point of \placeholder, as shown in the top of Figure \ref{fig:iSearch}.
The selection process is based on a Monte Carlo test, where we are given a timing constraint $TC$ and the search space iSpace.
We can prune the models whose minimum latency in the test fails to meet $TC$.
The feasible architectures will be sorted in terms of a weighted reward (will be introduced in Formula \ref{equ:reward}) in terms of the minimal latency and original accuracy.
Then, Top-K architectures will be selected as a starting point, where K is a user-defined variable.}



Now, iSearch gets into the second step to conduct the neural architecture search based on these selected models to make them meet the given timing constraint with \rtypos{high accuracy}.
iSpace tool will provide search spaces for iSearch, including the filter patterning $P$, channel cutting $C$, quantization $Q$, filter expansion $X$, and hardware design $H$.
All these search spaces are coupled with each other\rtypos{. In iSearch tool,} we develop a reinforcement learning based optimizer to simultaneously explore all these spaces.
Kindly note that other optimization techniques such as evolutionary algorithms \cite{real2017large} can be easily plugged into the \rtypos{iSearch} tool.
For better understanding, we will present the details of the optimizer at the end of this section.

\vspace{3pt}
\textit{iv) Problem definition:} 
\vspace{3pt}

Based on all the above definitions, we formally define the \rtypos{architecture-hardware co-search optimization problem} as follows: given a model zoo $M$, a specific FPGA $FP$, the accelerator design $D_i$ of model $A_i$ in $M$ on $FP$, a target timing constraint $T$, and the baseline accuracy $acc\_baseline$ we are going to determine:
\begin{itemize}
    \item $S$: \rtypos{selection of architectures from zoo $M$}, denoted as $A_0$;
    \item $P$,$C$,$X$,$Q$: \rtypos{tuning architecture hyperarameters of $A_0$};
    \item $H$: \rtypos{tuning hardware design hyperparameters on $D_0$}.
\end{itemize}
such that a new architecture $A_0^{\prime}$ \rtypos{with competitive accuracy over $acc\_baseline$ can be identified, while $A_0^{\prime}$ under hardware design $D_0^{\prime}$ can meet the timing constraint $T$.}







\vspace{5pt}
\noindent{\textbf{\emph{\raisebox{-1pt}{\large\ding{193}}}} \textbf{\textit{iSpace: An Integrated Search Space}}}
\vspace{5pt}

\revise{iSpace links the compression technique with the neural architecture search and hardware design.
In this work, we consider three model compression techniques: \textit{i)} pattern pruning; \textit{ii)} channel pruning; and \textit{iii)} quantization.
In the neural architecture search space, we consider \textit{iv)} filter expansion; for hardware design space, we mainly consider \textit{v)} communication bandwidth allocation and loop tiling, because FPGA accelerator design has typical templates which provides the above two kinds of hyperparameters in the design phase.
Kindly note that, \placeholder~is an open framework, {with} designers {having} the flexibility to modify or add new search parameters in terms of design needs.
{As an example}, dataflow and loop order can be further integrated into hardware design space when it comes to ASIC design {However, this} is out the scope of this work.
}

\begin{figure}[t]
  \centering
  \includegraphics[width=1\linewidth]{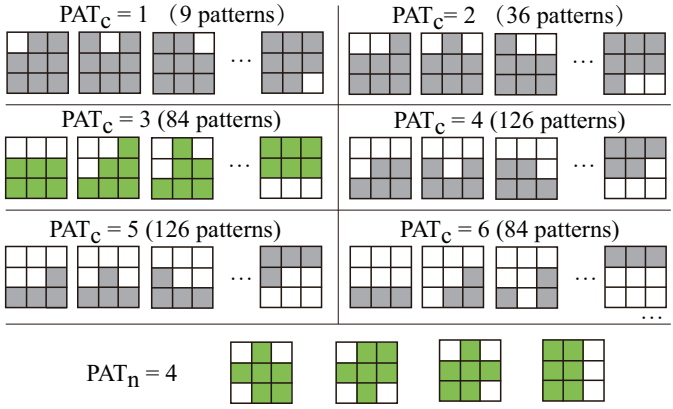}
  \vspace{-10pt}
  \caption{Patterns pruning for $3\times 3$ filters with different pruning categories ($PAT_c$); each big square represent a pattern, and each blank square in a pattern indicates the weights in the corresponding position to be pruned. The last row shows the number of patterns ($PAT_n$) and the specific patterns to be selected for pruning filters.}
  \label{fig:iSpace_Pattern}
\end{figure}


\textit{i) $P$: Pattern Pruning}
\vspace{3pt}

The first search space is the \rtypos{pattern pruning} space, which prunes the filter in the neural architecture $A$.
A pattern is defined as a mask matrix $Mat[x][y]$; $Mat[x][y]=0$ indicating that the weights at position $\langle x,y\rangle$ will be pruned, while $Mat[x][y]=1$ indicates that the weights will \rtypos{remain}.
According to the number of $0$ in $Mat[x][y]$, we can classify the pattern into different categories, and we use $PAT_c$ to indicate the number of $0$ in the pattern, as shown in Figure \ref{fig:iSpace_Pattern}.
Among all patterns, one category will be selected for pruning. 
Each pattern category is further composed of many patterns; for instance, there are 84 potential patterns in the category of $PAT_c=3$, as shown in Figure \ref{fig:iSpace_Pattern}.
For the hardware implementation, it simply cannot apply so many patterns \rtypos{as} this will \rtypos{results} in a large number of multiplexers in hardware implementation, \rtypos{making} the design inefficient.
In consequence, we will select a small number of patterns from the selected category, denoted as $PAT_n$.
Figure \ref{fig:iSpace_Pattern} gives the example of the \rtypos{pattern pruning} space  for $3\times 3$ filter, which selects the category of $PAT_c=3$ and \rtypos{applies} $PAT_n=4$ patterns \rtypos{among 84 candidates}. 

The selected patterns will be applied for a set of filters. 
As demonstrated in \cite{ma2019pconv}, by applying the Euclidean norm, we can specify one pattern for each kernel in a filter, i.e., the determination of $p(f_i)$ (see the definition in \textbf{\emph{\raisebox{-1pt}{\large\ding{192}}}} iSearch).
However, when \rtypos{implementing pattern pruning on hardware, the following two questions needing to be answered}: (1) How many kernels in a filter will be pruned by each type of pattern. (2) Whether all layers need to be pruned or  which layers will be pruned.
For the first question, it is related to the tiling design parameters. 
In a tile, if multiple types of patterns are applied, \rtypos{it will break} the execution pipeline and pattern pruning cannot improve performance at all.
This will be shown in \textbf{\emph{\raisebox{-1pt}{\large\ding{194}}}} iDesign.
For the second question, \rtypos{applying patterns for the layers whose performance bottleneck is at communication, it will not help in improving performance but may reduce accuracy}. 
Details will be illustrated in \textbf{\emph{\raisebox{-1pt}{\large\ding{195}}}} iDetect.

\vspace{3pt}
\textit{ii) $C$: Channel Pruning} 
\vspace{3pt}

\begin{figure}[t]
  \centering
  \includegraphics[width=1\linewidth]{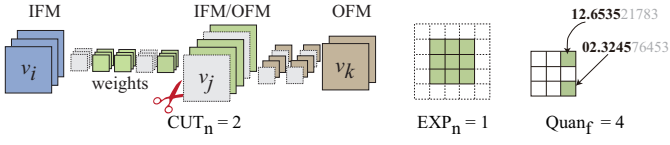}
  \vspace{-10pt}
  \caption{Three architecture search spaces: (left) channel cutting with ratio parameter $CUT_r$, (middle) kernel expansion with size parameter $EXP_s$, (right) weight quantization with fraction parameter $Quan_f$.}
  \label{fig:iSpace_others}
\end{figure}

\rtypos{Unlike \rtypos{pattern pruning}} \rtypos{that changes the structure}, the neural architecture will not be changed, with the channel pruning modifying the number of channels for a node $v_i\in V$ \rtypos{in} architecture $A$.
The left figure in Figure \ref{fig:iSpace_others} shows the channel pruning, where $CUT_n$ represents the number of channels to be cut off. \rtypos{We take $CUT_n=2$ in this example.} There are three consecutive nodes $v_i\rightarrow v_j \rightarrow v_k$, and we perform the channel \rtypos{pruning} on $v_j$.
In this figure, the grey channels in $v_j$ indicate the ones \rtypos{to be} cut off.
A ripple effect is taken to both filters of $f_{i\rightarrow j}$ and $f_{j\rightarrow k}$.
However, as the channel pruning may easily result in the accuracy drop since features are directly removed, we carefully \rtypos{formulate its search space for channel pruning only if the performance bottlenecks cannot be alleviated by other techniques (details in \textbf{\emph{\raisebox{-1pt}{\large\ding{195}}}} iDetect).}

\vspace{3pt}
\textit{iii) $Q$: Quantization}
\vspace{3pt}

Quantization is another model compression technique.
In general, the original model applies the data type of 32-bit floating-point, and we can convert it to the 16-bit fixed point without accuracy loss.
Such a fixed point representation is composed of two parts, the integer and fraction parts represented by $\langle I, F\rangle$.
For a given pre-trained architecture $A$, we can get the maximum and minimum parameters of one operator.
Then, we can analyze the number of bits required by integer part $I$.
Since the integer part is the most-significant bits, we will keep \rtypos{its bit-width}, and further squeeze the fraction part $F$ only, denoted as $Quan_f$ \rtypos{as shown in the right part of Figure \ref{fig:iSpace_others}}.
As will show in \textbf{\emph{\raisebox{-1pt}{\large\ding{195}}}} iDetect, not all layers \rtypos{need} to perform quantization, since it cannot alleviate specific types of 
performance bottlenecks.

\vspace{3pt}
\textit{iv) $E$: Filter Expansion}
\vspace{3pt}

The previous \rtypos{three search spaces belong to model compression; while filter expansion belongs to neural architecture search space.
This is motivated by the following two aspects}: (1) many state-of-the-art neural architectures identified by NAS contains larger sized filters, and (2) for specific layers, \rtypos{the increase of filter sizes will not add latency overhead}.
This will be shown in \textbf{\emph{\raisebox{-1pt}{\large\ding{195}}}} iDetect.
We define $EXP_n$ as the expansion factor on a filter, as shown in the middle \rtypos{part} of Figure \ref{fig:iSpace_others}.

\rtypos{Furthermore}, we have the following \rtypos{theorem} to guarantee that the accuracy will not be reduced by expanding the kernel.
\begin{theorem}
Given a pre-trained model $A=\langle V,E,r,c,ch,o,f,para,acc\rangle$, for any operator $o_i$ on edge $e_i$, the expansion on filter $f_i$ with factor $EXP_n$ will not decrease the accuracy, if the initial weights of the \rtypos{newly}  added weights on $f_i$ are set to 0, and \rtypos{$o_i$ is padded by $EXP_n$}.
\end{theorem}

The proof of the above property is straightforward, since \rtypos{all computations} remain the same when we increase the kernel size and padding with extra 0s.
With the guarantee of no accuracy loss, the expanded kernel makes it possible to increase accuracy after a \rtypos{fine-tuned} process.


\vspace{3pt}
\textit{v) $H$: Hardware Design Space}
\vspace{3pt}

Finally, after the modifications \rtypos{to}  architectures, the original hardware design identified by the optimization algorithms may not be the optimal one. 
In iSpace, we also provide flexibility to modify the hardware design and build the hardware design space.
In particular, according to the existing performance bottleneck, we create a \rtypos{search space} to adjust \rtypos{bandwidth-related design hyperparameters: $\langle I_b,O_b,W_b\rangle$}, and computation-related design hyperparameters, $\langle T_m, T_n, T_r,T_c\rangle$.

\vspace{5pt}
\noindent{\textbf{\emph{\raisebox{-1pt}{\large\ding{194}}}} \textbf{\textit{iDesign: Compression-Aware Performance Model}}}
\vspace{5pt}

\begin{figure}[t]
  \centering
  \includegraphics[width=1\linewidth]{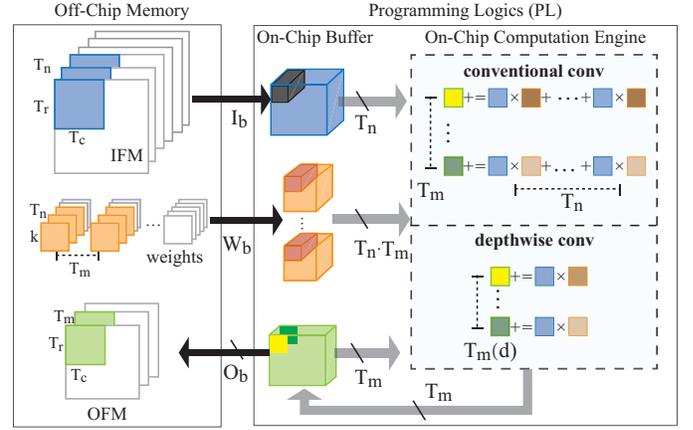}
  \vspace{-10pt}
  \caption{Illustration of accelerator architecture and design parameters: (left) off-chip memory to hold intermediate data and weights; (right) on-chip buffer and accelerator design.}
  \label{fig:iDesign}
\end{figure} 

Figure \ref{fig:iDesign} demonstrates the \rtypos{overview of system design}, where the left-hand part is the off-chip memory to hold IFM, OFM, and weight; while the right-hand part is the on-chip accelerator design that implements both conventional convolution and depthwise convolution using on-chip computing resource (e.g., DSPs).
In the accelerator \rtypos{design}, say conventional convolution, a set of multiplication-and-accumulation are computed in parallel.
\revise{Such a design has been using in many research works \cite{zhang2015optimizing,jiang2019achieving,jiang2018heterogeneous}; however, it still lacks a systematic model to efficiently support depthwise convolution and different compression techniques.
In the following text, we will first overview the performance model of the conventional convolution \cite{jiang2019achieving}, and then we revise the performance model to support depthwise convolution and compression.}



First, we introduce the computing accelerator part. Let $\mathbb{D}$ be the number of DSPs in the given FPGA, and $K$ be the size of the filter. 
\revise{As shown in the right-hand part in Figure \ref{fig:iDesign}, the conventional convolution involves $T_m\times T_n$ multiplication and additions (MAC).
For 16-bit data, each MAC needs one DSP.
In addition, to consume all data in on-chip buffers, it needs to repeat $K\cdot K\cdot Tr\cdot Tc$ times for computation; and the pipeline initial interval (II) is optimized to 1 cycle. Therefore, we have the following constraints on computing resources and latency.}
\begin{equation}\label{equ:dsp_fix}
T_m\times T_n\le \mathbb{D}
\end{equation}
\begin{equation}\label{equ:comp_lat}
tComp = K\cdot K\cdot Tr\cdot Tc\times 1
\end{equation}
where $tComp$ is the latency of computation for all data provided by the on-chip buffer.

\revise{Second, the size of the on-chip buffer is limited by $\mathbb{B}$. There are three types of data in communication: IFM, OFM, and weights. We need to determine the on-chip buffer size for each type of data, denoted as $bI$, $bO$, $bW$, which can be easily obtained from the left part in Figure \ref{fig:iDesign}.
Kindly note that the size of one on-chip buffer (BRAM) is limited, say 18K for ZCU102.
For the dimension of data that needs to be accessed in parallel (e.g., channels of IFM, i.e., $T_n$), they need to be placed in different BRAMs. 
Hence, the amount of data without a parallel requirement (e.g., $T_r$ and $T_c$ in IFM) is divided by 18K.
Finally, the size of the buffer is equal to 2 times tile size, where 2 indicates the double buffer utilized to hide communication by computation.
We have the following constraints.}
\begin{equation}\label{equ:bi}
bI = 2\times T_n\times \lceil T_r\cdot T_c\cdot bit_I/18K\rceil
\end{equation}
\begin{equation}
bO = 2\times T_m\times \lceil T_r\cdot T_c\cdot bit_O/18K\rceil
\end{equation}
\vspace{-10pt}
\begin{equation}\label{equ:bw}
bW = 2\times T_m\times T_n \times\lceil K\cdot K\cdot bit_W/18K\rceil
\end{equation}
\begin{equation}\label{equ:bram_all}
bI+bO+bW \le \mathbb{B}
\end{equation}
where $bit_I$, $bit_W$, $bit_O$ are the \rtypos{bit-width} of the data type used for IFM, weights, and OFM.

\revise{Third, based on the buffer size and the bandwidth (I\_b, W\_b, O\_b) allocated for each type of data buffer, we can get the communication latency ($tI_{mem}$, $tW_{mem}$, $tO_{mem}$) as follows.}
\begin{equation}\label{equ:tIFM}
tI_{mem} = \lceil T_n\cdot T_r\cdot T_c\cdot bit_I/I_b\rceil
\end{equation}
\begin{equation}\label{equ:tW}
tW_{mem} = \lceil T_m\cdot T_n\cdot K\cdot K\cdot bit_W/W_b\rceil
\end{equation}
\begin{equation}\label{equ:tOFM}
tO_{mem} = \lceil T_m\cdot T_r \cdot bit_O \cdot T_c/O_b\rceil
\end{equation}
\begin{equation}
(I_b+W_b+O_b) \le \mathbb{W}
\end{equation}
\revise{where $\mathbb{W}$ is the maximum bandwidth between off-chip memory and on-chip memory.}


Finally, based on the above formulations, we can derive the latency model. Let $M$, $N$, $R$, $C$ be the number of OFM channels, IFM channels, rows, and columns of the convolution layer.
We have the following models.
\begin{equation}\label{equ:lat1}
Lat_1 = \max\{tComp,tI_{mem},tW_{mem}\}
\end{equation}
\vspace{-10pt}
\begin{equation}\label{equ:lat2}
Lat_2 = \max\{\lceil \frac{N}{T_n}\rceil\cdot Lat_1,tO_{mem}\}
\end{equation}
\vspace{-6pt}
\begin{equation}\label{equ:Lat}
Lat = \lceil \frac{R}{T_r}\rceil\times \lceil \frac{C}{T_c}\rceil\times \lceil \frac{M}{T_m}\rceil\times Lat_2+ (tO_{mem}+Lat_1)
\end{equation}
Since \rtypos{OFM is} reused and stay in on-chip, \rtypos{it will be flushed} to off-chip memory when IFM and weights are loaded for $\lceil \frac{N}{T_n}\rceil$ times.
$Lat_1$ indicates the latency of computation, loading IFM, loading weights to be fired once, and $Lat_2$ indicates the latency of OFM to be flushed to off-chip memory.
Finally, for one layer, OFM is flushed to off-chip memory for $B\times \lceil \frac{R}{T_r}\rceil\times \lceil \frac{C}{T_c}\rceil\times \lceil \frac{M}{T_m}\rceil$ times, and we have the total latency $Lat$.

For the model of depthwise convolution, we only need to modify $T_m$ in the above formulas to be $T_m(d)$ and $Tn$ to be $1$.
Kindly note that we consider the real-time scenario where the batch size is 1, and therefore, the communication subsystem (including on-chip buffer model Formulas \ref{equ:bi} to \ref{equ:bram_all}, and off-chip memory access model Formulas \ref{equ:tIFM} to \ref{equ:tOFM}) of two types of convolutions are shared.
However, the accelerators are independent; therefore, we revise Formula \ref{equ:dsp_fix} as follows.
\begin{equation}\label{equ:dsp_fix_2}
T_m\times T_n + T_m(d)\le \mathbb{D}
\end{equation}

\vspace{5pt}
\noindent{\textbf{\emph{\raisebox{-1pt}{\large\ding{195}}}} \textbf{\textit{iDetect: Performance Bottleneck Detector}}}
\vspace{5pt}

Based on the iDesign, we have several observations for the techniques introduced in  \textbf{\emph{\raisebox{-1pt}{\large\ding{193}}}} iSpace, and we propose iDetect tool to analyze these search spaces in turn.
Kindly note that all operators in a neural architecture will be implemented on one board and reuse these resources.
Before \rtypos{discussing} each search space, we first present the following corollary to detect the performance bottleneck of a layer based on iDesign.
\begin{property}\label{Corollary1}
Given a layer and design parameters, we can detect the performance bottlenecks by considering $Lat_1$ and $Lat_2$ as follows:
\begin{itemize}[noitemsep,topsep=0pt,parsep=0pt,partopsep=0pt]
  \item $\bf{O}$: if $Lat_2$ is dominated by $tO_{mem}$, the performance bottleneck is on transmitting OFM data, otherwise,
  \item $\bf{I}$: if $Lat_1$ is dominated by $tI_{mem}$, the performance bottleneck is on transmitting IFM data,
  \item $\bf{W}$: if $Lat_1$ is dominated by $tW_{mem}$, the performance bottleneck is on transmitting weights,
  \item $\bf{C}$: if $Lat_1$ is dominated by $tComp$, we have fully utilized the involved computation resource.
\end{itemize}
\end{property}

\vspace{3pt}
\textit{i) Pattern pruning can reduce computation time}
\vspace{3pt}

\begin{figure}[t]
  \centering
  \includegraphics[width=1\linewidth]{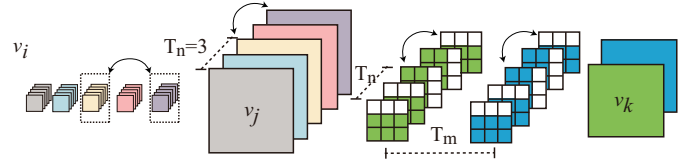}
  \vspace{-10pt}
  \caption{Reorder the input feature maps to make the \rtypos{pattern pruning} take effects in reducing the computation latency.}
  \label{fig:idesign_pattern}
\end{figure} 

Now, we are ready to answer the (1) question left in \textbf{\emph{\raisebox{-1pt}{\large\ding{193}}}} i): the number of \rtypos{kernels} pruned by each type of pattern is coupled with the titling factor $T_m$ and $T_n$.
As we can see from Figure \ref{fig:iDesign}, the data movement from on-chip weight buffer to \rtypos{the accelerator} is conducted in a pixel-wise way. As a result, it requires $K\times K$ iterations to \rtypos{traverse} the whole filter. To \rtypos{enable the effect of }, we need to make sure that \rtypos{all patterns} in one data tile are the same, as such we can skip these pruned weights in the outer loop to reduce the computation time.
In this way, we can modify Formula  \ref{equ:comp_lat} as follows.
\begin{equation}\label{equ:comp_lat_2}
tComp = (K\cdot K - PAT_n)\cdot Tr\cdot Tc
\end{equation}
where $PAT_n$ is the number of 0s in the pattern mask.

Next, since the pattern selection for kernels \rtypos{is based on the Euclidean norm}, it cannot guarantee all patterns for \rtypos{same type of data tiles}. 
We propose the input feature map reorder method to solve this problem.
As shown in Figure \ref{fig:idesign_pattern}, we can change the \rtypos{third and fifth} channels for the filter used in the operator $o_{j,k}$.
\rtypos{Correspondingly, we need to switch the} feature map in node $v_j$. \rtypos{It will also affect the} operator from $v_i$ to $v_j$, \rtypos{where we need to switch} the third and fourth filters.
In this way, we can make the pattern pruning take effects \rtypos{and reduce} the computation latency.

From Formulas \ref{equ:bw} and \ref{equ:tW}, it \rtypos{may appear} that pattern pruning can also reduce the on-chip buffer size and latency of loading weights. However, for buffer size, all layers reuse this buffer, and it cannot be specialized for one layer; while for loading weights, the pattern pruning will lead the loading procedure from sequential memory access to random access, \rtypos{as a result the latency may be even increased}. Hence, we will keep the sequential memory access to guarantee performance. 

\begin{property}
By applying the proposed reorder technique, pattern pruning can be employed to reduce the computing latency, but cannot reduce the latency of loading weights.
\end{property} 

\vspace{3pt}
\textit{ii) Channel pruning can conditionally reduce latency}
\vspace{3pt}

Channel pruning directly reduces the number of channels of feature maps in a node, and it \rtypos{can potentially} reduce the latency.
Let $Cut_n$ be the number of channels cut on the feature maps of node $v_i$.
When $v_i$ acts as the input feature maps for an operator, we need to modify \rtypos{Formula \ref{equ:lat2}} as follows:
\begin{equation}\label{equ:lat2_r}
Lat_2 = \max\{\lceil \frac{N-Cut_n}{T_n}\rceil\cdot Lat_1,tO_{mem}\}
\end{equation}
Then, when $v_i$ acts as the output feature maps for an operator, we revise \rtypos{Formula \ref{equ:Lat}} as follows:
\begin{equation}\label{equ:Lat_r}
Lat = \lceil \frac{R}{T_r}\rceil\times \lceil \frac{C}{T_c}\rceil\times \lceil \frac{M-Cut_n}{T_m}\rceil\times Lat_2+ (tO_{mem}+Lat_1)
\end{equation}

\begin{property}
\rtypos{Channel pruning can reduce the latency of a layer if and only if} (1) $\lceil\frac{M-Cut_n}{T_m}\rceil\le \lceil\frac{M}{T_m}\rceil$ or (2) $\lceil \frac{N-Cut_n}{T_n}\rceil<\lceil \frac{N}{T_n}\rceil$ and $Lat_2$ is not dominated by storing OFM data.
\end{property} 

The above property indicates that \rtypos{pruning} a small number of channels \rtypos{may not make an} impact. \rtypos{As such,} it guides the iSpace of channel pruning to take $T_m$ or $T_n$ as the step.



\vspace{3pt}
\textit{iii) Quantization can reduce latency of loading weights}
\vspace{3pt}

\rtypos{Quantization} is widely used in the neural network based FPGA implementations. It is demonstrated hybrid quantization can achieve good performance \cite{wang2019haq}, where weights in different layers have different bit-widths.
When we adopt such a hybrid approach, what benefits can be achieved?
From Formula \ref{equ:tW}, we can see that the quantization can take effects in reducing the latency of loading weights.
This can be implemented by composing multiple weights into one package.
\revise{As with computing latency, since the initial interval is already optimized to 1 cycle as shown in Formula \ref{equ:comp_lat}, the lower bit-width operations cannot further reduce clock cycles.
Lower bit-width can reduce the number of computing resources and have the potential to achieve high clock frequency.
However, when we consider an end-to-end implementation, the computing engine is shared by all layers.
Therefore, the layer with the largest bit-width dominates the design performance.}

\begin{property}
\revise{Quantization on a single layer can reduce latency of loading weights, but it may not reduce the computation latency if there exists another layer with larger bit-width.}
\end{property}

\vspace{5pt}
\noindent{\textbf{\emph{\raisebox{-1pt}{\large\ding{196}}}} \textbf{\textit{Optimizer: Search space exploration}}}
\vspace{5pt}

Finally, we introduce the RNN-based reinforcement learning optimizer employed in iSearch. 
As shown in the bottom part of Figure \ref{fig:iSearch}, \rtypos{an RNN} controller is designed based on the created design space by the \textbf{\emph{\raisebox{-1pt}{\large\ding{193}}}} iSpace tool.
\revise{Specifically, the controller is composed of a softmax classifier to predict hyperparameters for each search space in iSpace (e.g., quantization $Quan_f$ for a layer).
The predicted hyperparameters will identify a specific neural network and hardware design, which can derive a reward in terms of accuracy and latency.
The search process will optimize the controller by tuning its parameters $\theta_c$ to maximize the expectation of the reward.
A policy gradient method will be employed to update parameters $\theta_c$, aiming to predict better architectures over a series of episodes.}

\revise{In each episode, the predicted hyperparameters can be regarded as actions.
Based on the actions, the optimized neural architecture $A$ and hardware design $D$ can be derived.
In order to update the controller for the next episode, we need to compute the reward according to the following procedures: (1) calculate latency $lat$ of architecture $A$ on design $D$ by using the performance models proposed in \textbf{\emph{\raisebox{-1pt}{\large\ding{194}}}} iDesign and \textbf{\emph{\raisebox{-1pt}{\large\ding{195}}}} iDetect;
(2) verify whether timing constraint $T$ can be satisfied; if $lat > T$, we will directly calculate the reward without fine-tuning the model, otherwise, the reward is calculated based on accuracy and latency in the next step;
(3) fine-tune architecture $A$ to obtain accuracy $acc$ on a hold-out dataset; since the model is pre-trained, we do not need to train the model from scratch; instead, we fine-tune the model for a small number of data batches (not epochs), say $\beta=10$, to obtain $acc$.
Finally, the calculation of reward is based on the following formula:
\begin{equation}\label{equ:reward}
    R(acc,lat) = \alpha\times r\_acc + (1-\alpha)\times r\_lat
\end{equation}
where $\alpha$ is a scaling parameter to control with the search is for higher accuracy (i.e., larger $\alpha$) or lower latency (i.e., smaller $\alpha$).
If $lat>T$ indicating that the timing constraint cannot be satisfied, we have $r\_acc=-1$ and $r\_lat=T-lat$; otherwise, we normalize $r\_acc$ and $r\_lat$ to the range from -1 to 1, as follows: $r\_acc=\frac{acc-A\_min}{A\_ori-A\_min}\times2-1$ and $r\_lat=\frac{T-lat}{T-T\_min}\times2-1$, where $A\_ori$ is the original accuracy of backbone architecture; $T$ is the timing constraint; $A\_min$ and $T\_min$ are the lower bounds on accuracy and latency, which are involved for a better normalization. 
}


Based on the reward function, the optimizer will iteratively work in two steps.
First, the controller predicts a sample, and gets its reward $R$.
Then, the Monte Carlo policy gradient algorithm \cite{williams1992simple} is employed to update the controller:
\begin{equation}
    \nabla J(\theta) = \frac{1}{m}\sum_{k=1}^{m}\sum_{t=1}^{T}\gamma^{T-t}\nabla_{\theta}\log \pi_{\theta} (a_{t}|a_{(t-1):1})(R_{k}-b)
\end{equation}
where $m$ is the batch size and $T$ is the number of steps in each episode. Rewards are discounted at every step by
an exponential factor $\gamma$ and baseline $b$ is the average exponential moving of rewards.



\section{Experimental Results}

\begin{table}[t]
  \centering
  \tabcolsep 8.3pt
  \renewcommand\arraystretch{1.3}
  \caption{\revise{Model selection based on iDesign and iDetect to conduct 100 Monte Carlo Test in iSpace, with the timing constraint of $T\le 5ms$.}}
    \begin{tabular}{cccccc}
    \toprule
    \multirow{2}{*}{Models} & Original   & \multicolumn{4}{c}{Latency (ms) in Monte Carlo Tests} \\
\cline{3-6}          & Latency (ms) & Min   & Sat.  & Max   & Ave. \\
    \midrule
    ProxylessNAS & 5.83  & 4.47  & \checkmark     & 5.95  & 5.08 \\
    MnasNet & 5.94  & 4.68  & \checkmark     & 7.88  & 5.21 \\
    ResNet & 6.27  & 4.31  & \checkmark     & 5.9   & 5.05 \\
    MobileNet & 6.29  & \textbf{5.07} & $\times$     & 7.88  & 6.2 \\
    FBNet & 7.37   & \textbf{5.85}  & $\times$     & 7.82  & 6.66 \\
    \bottomrule
    \end{tabular}%
  \label{tab:model_selection}%
\end{table}%

\begin{table*}[htbp]
  \centering
  \tabcolsep 6.5pt
  \renewcommand\arraystretch{1.5}
  \caption{On ImageNet, comparison of the state-of-the-art neural architectures with timing constraints of $5ms$.}
    \begin{tabular}{ccccccccccc}
    \toprule
    Model & Type  & Latency & Sat.  & Param. (\#) & Param. (S) & Top-1 & Top-5 & Top-1 Imp. & Top-5 Imp. & GPU Time \\
    \midrule
    AlexNet & manually & 2.02  & \checkmark     & 61.1M & 122.20MB & 56.52\% & 79.07\% & -     & -     & - \\
    \textit{MnasNet 0.5 *} & \textit{auto} & \textit{3.99} & \textit{\checkmark} & \textit{2.22M} & \textit{4.44MB} & \textit{67.60\%} & \textit{87.50\%} & \textit{-} & \textit{-} & \textit{40,000H} \\
    SqueezeNet 1.0 & manually & 4.76  & \checkmark     & 1.25M & 2.50MB & 58.09\% & 80.42\% & -     & -     & - \\
    \midrule
    ProxylessNAS & auto  & 5.83  & $\times$     & 4.08M & 8.16MB & 74.59\% & 92.20\% & -     & -     & 200H \\
    MnasNet & auto  & 5.94  & $\times$     & 4.38M & 8.77MB & 73.46\% & 91.51\% & -     & -     & 40,000H \\
    Resnet & manually & 6.27  & $\times$     & 11.69M & 23.38MB & 69.76\% & 89.08\% & -     & -     & - \\
    \revise{Co-Exploration \cite{jiang2020hardware}} & \revise{auto}  & - & - & \revise{-} & \revise{-} & \revise{70.42\%} & \revise{90.53\%} & \revise{-} & \revise{-} & \revise{266H} \\
    \midrule
    \textbf{\placeholder-Resnet(4ms)} & \textbf{auto} & \textbf{4.00} & \textbf{\checkmark} & \textbf{10.99M} & \textbf{17.49MB} & \textbf{68.27\%} & \textbf{88.21\%} & \textbf{0.67\%} & \textbf{0.71\%} & \textbf{2H22M} \\
    \textbf{\placeholder-Resnet} & \textbf{auto} & \textbf{4.22} & \textbf{\checkmark} & \textbf{11.19M} & \textbf{17.90MB} & \textbf{69.14\%} & \textbf{88.83\%} & \textbf{1.54\%} & \textbf{1.33\%} & \textbf{2H01M} \\
    \textbf{\placeholder-ProxylessNAS} & \textbf{auto} & \textbf{4.86} & \textbf{\checkmark} & \textbf{4.38M} & \textbf{8.31MB} & \textbf{73.39\%} & \textbf{91.47\%} & \textbf{5.79\%} & \textbf{3.97\%} & \textbf{2H37M} \\
    \textbf{\placeholder-Mnasnet} & \textbf{auto} & \textbf{4.99} & \textbf{\checkmark} & \textbf{4.07M} & \textbf{6.56MB} & \textbf{73.24\%} & \textbf{91.37\%} & \textbf{5.64\%} & \textbf{3.87\%} & \textbf{1H50M} \\
    \midrule
    \multicolumn{11}{l}{``$^{*}$'': baseline; ``$auto$ $\&$ $manually$'': the model identified by NAS or human experts; ``$\times$ $\&$ \checkmark'': violate or meet timing constraints.} \\
    \multicolumn{11}{l}{\revise{Results of Co-Exploration is derived from \cite{jiang2020hardware}, since the pre-trained model is not provided; The latency is not reported since it uses different hardware.}} \\
    \bottomrule
    \end{tabular}%

  \label{tab:comp}%
\end{table*}%

\revise{The proposed \placeholder~is evaluated on commonly used datasets, ImageNet \cite{deng2009imagenet} and CIFAR-10 with Xilinx ZCU102 board.}
In the following texts, we will first introduce the experimental setup. 
Then, we will compare \placeholder~with the state-of-the-art models to show that \placeholder~can achieve up to 5.79\% top-1 accuracy gain with the same timing constraint.
Next, we will visualize the results explored by \placeholder, followed by the design space exploration results to demonstrate the importance of co-exploring all design spaces in iSpace.
\revise{Finally, we report results and detailed analysis on CIFAR-10, showing that \placeholder~can achieve consistent improvement on different datasets.}

\subsection{\revise{Experimental setup}}\label{sec:exp_set}



\textit{Model Zoo.} For ImageNet dataset, we apply all models in torchvision, including AlexNet, VGGNet, ResNet, MobileNet-v2, Mnasnet, etc., as shown in Figure \ref{fig:existingnets}.
We also include the FBNet \cite{wu2019fbnet} and ProxylessNAS \cite{cai2018proxylessnas} for comparison.
\revise{In the experiments, we select a set of models to be optimized.
According to iDesign and iDetect, we run Monte Carlo Tests to get statistic latency for 100 solutions in iSpace, as shown in Table \ref{tab:model_selection}. We prune the models whose minimum latency cannot satisfy the timing constraints, say $T\le 5$ in our settings.
Kindly note that the maximum latency can be larger than the original model latency, because we change the hardware configuration during the search, which may reduce bandwidth for the data movement which is the performance bottleneck.
Based on the results in Table \ref{tab:model_selection}, we select ProxylessNAS (mobile), MnasNet 1.0 (depth multiplier of 1.0), and Resnet-18 (with 18 layers) \cite{he2016deep} for optimization.
We denote them as ProxylessNAS, Mnasnet, and Resnet, respectively.}


\revise{For CIFAR-10 dataset, we collect the 4 sets of pretrained models, including ResNet-18 \cite{he2016deep}, DenseNet-121 \cite{huang2017densely}, MobileNet-v2 \cite{sandler2018mobilenetv2}, and BiT \cite{Kolesnikov2019BigT},
among which BiT achieves the state-of-the-art accuracy on CIFAR-10 dataset, which is built on top of existing neural networks.
In our experiments, with the hardware performance consideration, we select the ResNet-50 based version for BiT, which provides a baseline with the accuracy of 97.07\% and latency of 6.88ms.
For a better presentation, we denote the above models as ResNet, DenseNet, MobileNet, and BiTNet, respectively.
}

\textit{iSearch.} 
\revise{In iSearch component, we first need to determine parameters $\alpha$ and $\beta$.}
We set $\alpha=0.7$ to generate the reward as shown in Formula \ref{equ:reward}, and set the number of batch size $\beta=10$ to be used in \rtypos{the fine-tune phase}.
\revise{Furthermore, we will investigate the effects on performance made by different configurations of $\alpha$ and $\beta$ on CIFAR-10.}
Second, we need to set the number of episodes for reinforcement learning; here, we set the maximum episode to be 2,000 which can guarantee the convergence of the controller.
After running iSearch, we can obtain a set of architectures, and we will select the best architectures, i.e., the architecture with the highest accuracy under the given timing constraints. 

\textit{iSpace.} A new module that supports pattern pruning, channel pruning, filter expansion, and quantization \rtypos{is} implemented in \rtypos{Pytorch by overriding} the existing Conv2d module. During the iSearch process, the module can be customized for each layer in terms of the searched parameters, and automatically integrated into the model with the original weights.

\textit{iDesign.} We apply Xilinx ZCU102 board with XCZU9EG chip as the implementation hardware, which is composed of 600K logic cells, 32.1Mb on-chip buffers, 2,520 DSPs.
For data movement between on-chip and off-chip memory, there are 4 HP ports with the bandwidth of 128 bits for each.

\subsection{\revise{Results on ImageNet}}
\textbf{\noindent {\textit{i)} Comparison with \placeholder}}


Table \ref{tab:comp} reports the comparison results of \placeholder~with the existing state-of-the-art models. 
In the table, the column ``Type'' shows whether the model is identified by NAS or manually designed.
The column ``Sat.'' shows \rtypos{whether the model satisfies} the timing constraint of $5ms$.
Columns ``Param. (\#)'' and ``Param. (S)'' reports the \rtypos{number of parameters} and the size of \rtypos{parameters}, respectively.
Columns ``Top-1'', ``Top5'', ``Top-1 Imp.'', and ``Top-5 Imp.'' are model accuracy and accuracy gain to the baseline model on ImageNet.
Column ``GPU Hours'' shows the cost to identify the model for all models identified by NAS.
Finally, the rows marked as bold are models identified by the proposed \placeholder.

From the results in Table \ref{tab:comp}, we have three important observations: (1) Directly plugging the existing models onto the target FPGA board will easily result in the latency to be violated; while the proposed \placeholder~can guarantee to find the architectures to meet the latency constraints, meanwhile achieving high accuracy.
(2) For the existing models that can directly satisfy the timing constraints, the highest top-1 accuracy and top-5 accuracy are merely 67.60\% and 87.50\%. \rtypos{In comparison}, \placeholder~can achieve 5.79\% and 3.97\% accuracy gain with 73.39\% for top-1 and 91.47\% for top-5. 
(3) The cost of the existing neural architecture search is extremely high, which is at least 200 GPU hours. In comparison, the proposed \placeholder only takes less than 3 GPU hours to identify the model. 
\revise{Furthermore, compared with the existing co-exploration method, the search time can be significantly reduced from 266 GPU hours with 2.97\% Top-1 accuracy gain. 
All these observations clearly demonstrate the superiority of \placeholder~to obtain solutions with high accuracy and low search cost.}

\begin{figure}[t]
  \centering
  \includegraphics[width=1\linewidth]{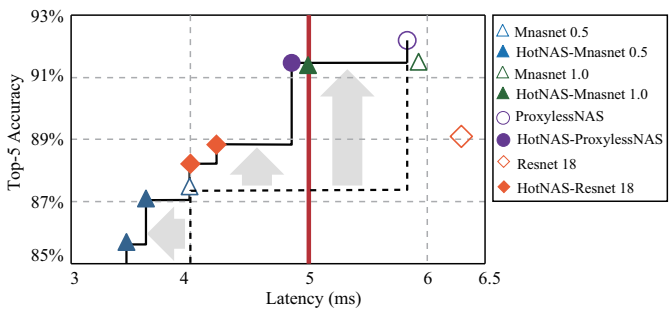}
  \caption{Pushing forward the Pareto frontier between latency and top-5 accuracy from the one built by the existing models to that by \placeholder under the timing constraint of 5ms.}
  \label{fig:exp_comp}
\end{figure} 

Besides, from the results, we can see that \placeholder performs good at \rtypos{reducing} the latency while \rtypos{maintaining high accuracy}.
For Resnet18, \placeholder can reduce the latency from 6.27ms to 4.22ms with 32.70\% reduction, while the top-5 accuracy loss is merely 0.25\%; for ProxylessNAS, the latency reduction is 16.64\% with only 0.53\% top-5 accuracy loss; for Resnet18, these figures are 15.9\% and 0.14\%.
We will have a detailed and visualized analysis in the latency reduction later in this section.
A further observation from the above result is that the manually designed Resnet18 can achieve larger reductions in latency than the automatically identified ones.
This is reasonable since the automatically designed architectures have already be used for optimizing for other platforms, while manually designed architectures may have more redundant parameters. This can also be observed by the reduction in both the number of parameters and the size of parameters.

Figure \ref{fig:exp_comp} \rtypos{further shows} the comparison of Pareto frontiers built by the existing models and \placeholder. In this figure, the x-axis and y-axis represent the latency and accuracy, respectively. The red line stands for the timing constraints. The solid points are solutions identified by \placeholder, while the hollow ones are the existing models.
The arrows in this figure clearly demonstrate that \placeholder~can significantly push forward the Pareto frontier between accuracy and latency in two directions: (1) vertical direction: improving accuracy; (2) horizon direction: reducing latency.
The results in this figure again demonstrate the efficiency and effectiveness of the proposed \placeholder.

\begin{table}[t]
  \centering
  \tabcolsep 5pt
  \renewcommand\arraystretch{1.3}
  \caption{Solution visualization, using the modified layers and latency reduction in \placeholder-Resnet18 as an example}
    \begin{tabular}{ccccc}
    \toprule
    Layers/HW & iDetect & iSpace & Exploration Results & Red. (ms) \\
    \midrule
    layer1[0].conv1 & \multirow{7}[2]{*}{C} &  \multirow{7}[2]{*}{Pattern} & PATr=3, PATn=4 & \multirow{7}[2]{*}{0.57} \\
    layer1[0].conv2 &       & & \multirow{6}[1]{*}{\raisebox{-0.8\totalheight}{\includegraphics[width=0.9in]{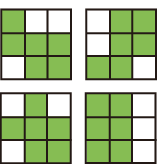}}} &  \\
    layer1[1].conv1 &       &&       &  \\
    layer1[1].conv2 &       &&       &  \\
    layer2[0].conv2 &       &&       &  \\
    layer2[1].conv1 &       &&       &  \\
    layer2[1].conv2 &       &&       &  \\
    \midrule
    layer4[0].conv1 &   \multirow{2}[2]{*}{I} & \multirow{2}[2]{*}{Channel}    & $512\rightarrow480$ & \multirow{2}[2]{*}{0.15}  \\
    layer4[1].conv1 &       && $512\rightarrow496$ &  \\
    \midrule
    layer4[0].conv1 & -  & \multirow{4}[2]{*}{Quant.} & \multirow{4}[2]{*}{$[1,15]\rightarrow[1,7]$} & \multirow{4}[2]{*}{1.01} \\
    layer4[1].conv1 & -      &&       &  \\
    layer4[0].conv2 &  \multirow{2}[2]{*}{W}    & &       &  \\
    layer4[1].conv2 &       &&       &  \\
    \midrule
    $I_b$    & \multirow{2}[2]{*}{-} & \multirow{2}[2]{*}{HW} & $18\rightarrow20$& \multirow{2}[2]{*}{0.32} \\
    $W_b$   & &&  $6\rightarrow5$  & \\
    \midrule
    \multicolumn{4}{c}{Total} & \textbf{2.05} \\
    \bottomrule
    \end{tabular}%
  \label{tab:exp_visu}%
\end{table}%

\textbf{\noindent{\textit{ii)} Results visualization}}

Table \ref{tab:exp_visu} shows the visualization results of  \placeholder-Resnet18.
For other resultant architectures, they have similar results, but the model is too large to demonstrate\footnote{This project will be open-source after the blind review, and all models with fine-tuned parameters can be accessed online.}.
In this table, column ``iDetect'' shows the performance bottleneck with the original design detected by \placeholder, and column ``iSpace'' shows the built search spaces for these the corresponding layers.
The column \rtypos{``exploration results''} show the detailed changes from the original architecture to the resultant model.
Finally, \rtypos{the column ``Red.''} shows the latency reduction contributed by each search space.

It is clearly shown in this table that the proposed \placeholder~can identify different types of performance bottleneck in the architecture, and apply the matched techniques to alleviate the performance bottlenecks.
Specifically, pattern pruning identifies 4 patterns in pattern category $PAT_r=3$, and achieves 0.57ms latency reduction.
Channel pruning, quantization, and hardware modifications \rtypos{achieve a reduction} of 0.15ms, 1.01ms, and 0.32ms, respectively.
As a whole, the reduction is 2.05ms, from 6.27ms to 4.22ms, as shown in Table \ref{tab:comp}.
Kindly note that since the latency of loading IFM and loading weights are quite close for layer 4, iSpace creates search spaces for both channel pruning and quantization.

\textbf{\noindent{\textit{iii)} No space in iSpace can be dispensed}}

\begin{figure}[t]
  \centering
  \includegraphics[width=1\linewidth]{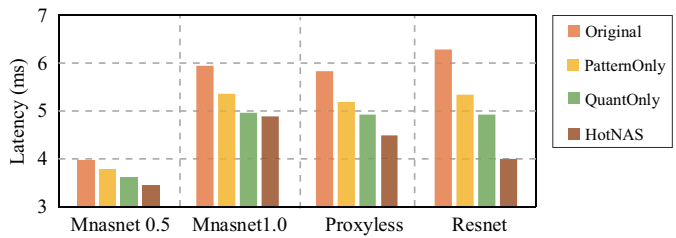}
  \caption{Comparison results in the latency reduction among three techniques using the original architecture as baseline: (1) PatternOnly: apply pattern pruning \cite{ma2019pconv}, (2) QuantOnly: apply hybrid quantization \cite{wang2019haq}, (3) \placeholder.}
  \label{fig:exp_oneN4all}
\end{figure} 

There are a lot of existing techniques \rtypos{that focus on devising} a specific technique for model compression. We compare with \rtypos{the two most effective methods} using pattern pruning only \cite{ma2019pconv}, denoted by PatternOnly; and hybrid quantization \cite{ma2019pconv}, denoted by QuantOnly. However, as discussed in this iDetect, no technique can cover all kinds of performance bottlenecks.
Results in Figure \ref{fig:exp_oneN4all} \rtypos{verify} this claim.
Kindly note that the hardware space is kept for all techniques for a fair comparison.
In this figure, the x-axis is the backbone architecture, and the y-axis is the latency that can be achieved with the same accuracy constraint.
The baseline is the original neural architecture without optimization.

Results in Figure \ref{fig:exp_oneN4all} clearly demonstrate that without fully considering the performance bottleneck and apply only one technique for optimization will lead to inferior solutions.
Taking Resnet as an example, PatternOnly can reduce the time from 6.27ms to 5.34ms, and QuantOnly can further \rtypos{reduce} it to 4.92ms.
By a full consideration of all kinds of bottlenecks, \placeholder~can achieve the architecture with 4ms, which achieves 25.09\% and 18.71\% latency reductions compared with PatternOnly and QuantOnly, respectively.
Results form this group of experiments emphasize the needs of an automatic tool to analyze the model, detect the performance bottlenecks, and alleviate each kind of bottlenecks using the correct technique.

\begin{figure}[t]
  \centering
  \includegraphics[width=1\linewidth]{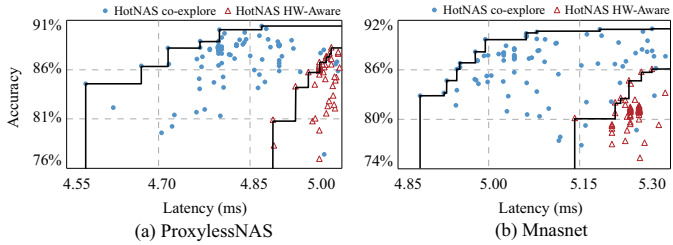}
  \vspace{-10pt}
  \caption{Illustration of importance of Co-Exploration.}
  \label{fig:co_explore}
\end{figure} 

In addition to the compression techniques, the hardware exploration is also important, this is shown from the results in Figure \ref{fig:co_explore}.
We compare the exploration results of \placeholder with the consideration of hardware modification space (``\placeholder~w/ HW'') and results of that without hardware modification (``\placeholder~w/o HW'').
Figures \ref{fig:co_explore}(a) and \ref{fig:co_explore}(b) demonstrate the results for ProxylessNAS and Mnasnet, respectively.
In these figures, the x-axis and y-axis are the latency and top-5 accuracy, respectively.
Each blue dot represents a solution explored by ``\placeholder~w/ HW'', while each red triangle represents that explored by ``\placeholder~w/o HW''.
The solutions with either higher accuracy or lower latency among each group of results \rtypos{form} the Pareto frontiers.

As we can see from Figure \ref{fig:co_explore}, the proposed \placeholder~can significantly push forward the Pareto frontiers.
Specifically, for ProxylessNAS, the smallest latency of solutions explored by ``\placeholder~w/ HW'' is 4.58ms, while that of ``\placeholder~w/o HW'' is 4.89ms; for accuracy, ``\placeholder~w/ HW'' can achieve 2.15\% accuracy gain with latency reduction of 0.11ms.
These results emphasize the importance of conducting co-exploration in neural architecture search.





\subsection{\revise{Results on CIFAR-10}}
\textbf{\noindent{\textit{i).} Pushing forward accuracy-latency Pareto frontier}}

\begin{figure}[t]
  \centering
  \includegraphics[width=1\linewidth]{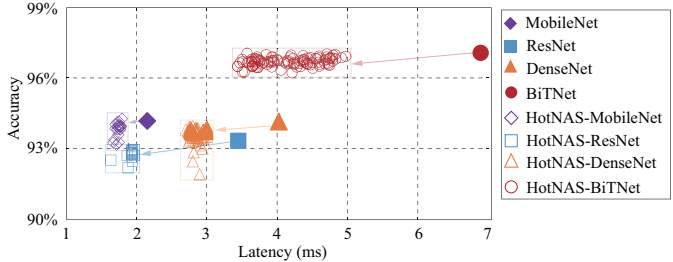}
  \caption{\revise{On CIFAR-10, exploration results of \placeholder~for 4 pre-trained models with different target latency constraints.}}
  \label{fig:cifarRes}
\end{figure}

\revise{\placeholder~can consistently push forward the accuracy-latency Pareto frontier for different datasets.
On CIFAR-10 dataset, we achieve similar results as ImageNet dataset.
Figure \ref{fig:cifarRes} reports the exploration results for four pre-trained models.
In this figure, x-axis and y-axis stand for latency and accuracy, respectively.
The solid shapes represent the performance baseline models, while others represent the explored results.
In order to evaluate the scalability of proposed \placeholder, unlike the previous experiment applying a uniform target latency, we set an individual target latency for each model according to the baseline latency.
Specifically, the target latency constraints are 2ms, 3ms, 1.8ms, and 5 ms for ResNet, DenseNet, MobileNet, BiTNet, respectively.}

\revise{Results in this figure show that all architectures identified by \placeholder~can satisfy latency constraint, while achieving similar accuracy with the baseline architectures.
As a result, the accuracy-latency Pareto frontier can be significantly pushed forward, likewise that for ImageNet.
Another interesting observation is that BiTNet has a loose timing constraint, and \placeholder~can find a wider range of results with high accuracy.}

\revise{Table \ref{tab:cifar_res} reports the detailed comparison of the best architectures identified by \placeholder~over the baseline model.
The best architecture is selected based on the architectures with the highest accuracy while satisfying the timing constraint.
Then, we fine-tune the selected architecture for 10 epochs to obtain the final accuracy.
The accuracy and latency for the original model and the one identified by \placeholder~are reported in Columns ``baseline'' and ``\placeholder'' under Columns ``Accuracy'' and ``Latency''.
}

\revise{From Table \ref{tab:cifar_res}, it is clear to see that \placeholder~can efficiently reduce the latency which achieving accuracy gain on CIFAR-10 dataset.
Specifically, for ResNet, \placeholder~identifies the solution with 43.90\% latency reduction and 0.03\% accuracy gain; these figures are 28.55\% and 0.05\% for DenseNet; 16.74\% and 0.10\% for MobileNet; 48.26\% and 0.06\% for BitNet.
The above results demonstrate the efficiency and effectiveness of \placeholder.
}


\begin{table}[t]
  \centering
  \tabcolsep 4pt
  \renewcommand\arraystretch{1.3}
  \caption{\revise{On CIFAR-10, comparison of the baseline models and the best solutions explored by \placeholder~after fine-tuning}}
    \begin{tabular}{ccccccc}
    \toprule
    \multirow{2}{*}{Model} & \multicolumn{3}{c}{Accuracy} & \multicolumn{3}{c}{Latency (ms)}  \\
\cline{2-7}    
\multicolumn{1}{c}{} & \multicolumn{1}{c}{baseline} & \multicolumn{1}{c}{\textbf{HotNAS}} & \multicolumn{1}{c}{\textbf{comp.}} & \multicolumn{1}{c}{baseline} & \multicolumn{1}{c}{\textbf{HotNAS}} & \textbf{impr.} \\
    \midrule
    ResNet & 93.33\% & \textbf{93.36\%} & \textbf{+0.03\%} & 3.44  & \textbf{1.93} & \textbf{43.90\%}  \\
    DenseNet & 94.14\% & \textbf{94.19\%} & \textbf{+0.05\%} & 4.01  & \textbf{2.87} & \textbf{28.55\%}  \\
    MobileNet & 94.17\% & \textbf{94.27\%} & \textbf{+0.10\%} & 2.14  & \textbf{1.79} & \textbf{16.74\%} \\
    BiTNet & 97.07\% & \textbf{97.13\%} & \textbf{+0.06\%} & 6.88  & \textbf{3.56} & \textbf{48.26\%}  \\
    \bottomrule
    \end{tabular}%
  \label{tab:cifar_res}%
\end{table}%

\textbf{\noindent{\textit{ii).} Exploration with different configurations}}

\revise{There are two hyperparameters in the RNN-based optimizer: $\beta$ for the batch size of fine-tuning in the search process; $\alpha$ for the weights in the reward formulation.
In the following, we will test different settings on both.}

\begin{table}[t]
  \centering
  \tabcolsep 2pt
  \renewcommand\arraystretch{1.3}
  \caption{\revise{On CIFAR-10, comparison of different setting in \placeholder~on fine-tine batch size $\beta$ during the search process; $\beta=195$ for 1-epoch-search and $\beta=10$ for fast-search}}
    \begin{tabular}{ccccccc}
    \toprule
    \multirow{2}{*}{Model} & \multicolumn{3}{c}{1-epoch-search} & \multicolumn{3}{c}{fast-search} \\
    \cline{2-7}
          & Accuracy & Latency & GPU Time & Accuracy & Latency & GPU Time \\
    \midrule
    ResNet & 93.36\% & 1.93  & 7M21S & 92.74\% & 1.84  & \textbf{3M26S} \\
    DenseNet & 94.08\% & 2.79  & 55M26S & \textbf{94.19\%} & 2.87  & \textbf{12M04S} \\
    MobileNet & 94.27\% & 1.79  & 20M15S & 94.21\% & 1.79  & \textbf{4M26S} \\
    BiTNet & 97.13\% & 3.56  & 2H20M & 97.04\% & 3.84  & \textbf{18M44S} \\
    \bottomrule
    \end{tabular}%
  \label{tab:exp_beta}%
\end{table}%

\revise{First, we apply two settings on $\beta$: (1) $\beta=195$ for ``1-epoch-search'' which will fine-tune the identified architecture using the whole training set;
(2) $\beta=10$ for ``fast-search'' which only use a portion of dataset as in ImageNet experiments;
Table \ref{tab:exp_beta} reports the results. 
We can see that fast-search can achieve competitive accuracy compare with 1-epoch-search; in particular, for DenseNet, fast-search achieves 0.11\% higher accuracy.
In addition, for all models, fast-search can find solutions in 20 minutes.
These results demonstrate the efficiency of \placeholder.}

\begin{figure}[t]
  \centering
  \includegraphics[width=1\linewidth]{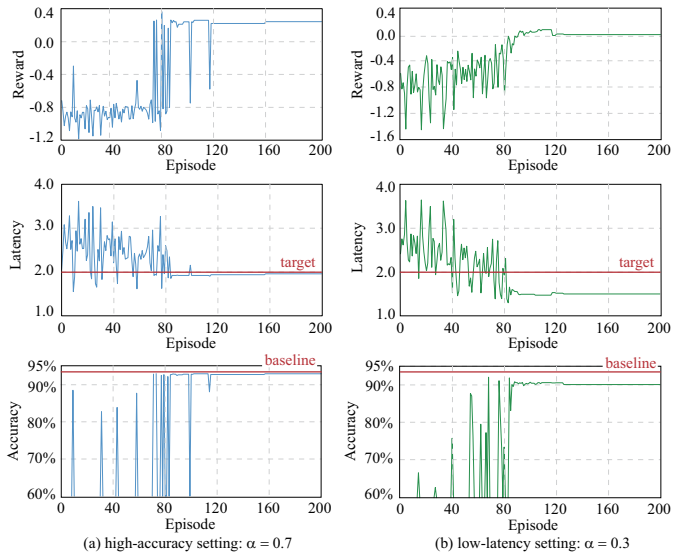}
  \caption{\revise{On CIFAR-10, comparison of different settings on scaling parameters $\alpha$ in optimizing ResNet.}}
  \label{fig:cifar_config}
\end{figure}

\revise{Figure \ref{fig:cifar_config} reports the results of two settings on $\alpha$: (1) $\alpha=0.7$ aiming at higher accuracy as shown in Figure \ref{fig:cifar_config}(a), (2) $\alpha=0.3$ aiming at lower latency as shown in Figure \ref{fig:cifar_config}(b).
We report the comparison results of ResNet on CIFAR-10, and we observe similar results for the other models and datasets.
In the experiments, we trace the reward, latency, and accuracy of solutions in each episode, which are reported in Figure \ref{fig:cifar_config}(a) and (b), respectively.
For latency, the red line shows the target latency, i.e., $\le 2.0ms$; while for accuracy, the red  line shows the baseline accuracy, which is 93.33\%.}

\revise{From Figure \ref{fig:cifar_config}, we can see that the search processes are converged after 160 and 120 episodes for the high-accuracy setting and the low-latency setting, respectively.
At the convergence, the latency of solutions identified by low-latency setting is lower than the high-accuracy setting; more interesting, the high-accuracy setting finds solutions with latency near to the threshold $2ms$.
For accuracy, we can see that high-accuracy setting finds solutions with higher accuracy, which is almost the same with the baseline accuracy. As shown in the results in Table \ref{tab:cifar_res}, after a fine-tuned process, the accuracy can even higher than the baseline.
One more thing noted by the accuracy results is that there are several episodes having no accuracy. This is because the latency cannot be satisfied, and we terminate the training procedure to accelerate the search process.}

\revise{All the above results show that \placeholder~provides flexibility for designers to better optimize neural architecture and hardware design according to their varied demands.}


\section{Conclusion}
In this work, we identify the last mile problem in current neural architecture search and hardware accelerator design and propose the \placeholder~toolset to solve the problem.
Instead of search architectures from scratch, we propose to stand on the shoulders of the existing models to conduct an incremental hardware-aware neural architecture search. 
In \placeholder, four components \rtypos{work} collaboratively to (1) identify the hardware performance bottleneck by iDetect, (2) build search spaces iSpace in terms of results from iDetect, (3) co-design the neural architecture and hardware accelerator by iSearch with the performance model \rtypos{provided} by iDesign.
Experimental results on ImageNet dataset demonstrate that \placeholder~can guarantee the resultant system to meet timing specifications, while achieving over 5.6\% top-1 and over 3.8\% top-5 accuracy gain, compared with the state-of-the-art models.


%



\bibliographystyle{ieeetr}
\bibliography{ref}

\end{document}